\documentclass[10pt,twocolumn,letterpaper]{article}

\usepackage{wacv}              

\usepackage{graphicx}
\usepackage{booktabs}
\usepackage{amsmath}
\usepackage{amssymb}

\usepackage{times}
\usepackage{epsfig}
\usepackage{comment}
\usepackage{epigraph}
\usepackage{multirow}
\usepackage{hhline}
\usepackage{makecell}
\usepackage{bm}
\usepackage{subcaption}
\usepackage{soul}
\usepackage{array}
\usepackage{color,colortbl}
\usepackage{xcolor}

\makeatletter
\def\blfootnote{\gdef\@thefnmark{}\@footnotetext}
\makeatother

\definecolor{beaublue}{rgb}{0.74, 0.83, 0.9}

\newcommand{\argmin}{\operatornamewithlimits{argmin}}

\definecolor{wacvblue}{rgb}{0.21,0.49,0.74}
\usepackage[pagebackref,breaklinks,colorlinks,allcolors=wacvblue]{hyperref}

\def\wacvPaperID{00042} 
\def\confName{WACV}
\def\confYear{2026}

\title{Procedure Learning via Regularized Gromov-Wasserstein Optimal Transport}

\author{Syed Ahmed Mahmood$^\dagger$~~~~~Ali Shah Ali$^\dagger$~~~~~Umer Ahmed$^\dagger$~~~~~Fawad Javed Fateh$^\dagger$\\M. Zeeshan Zia~~~~~Quoc-Huy Tran\\
\\
Retrocausal, Inc., Redmond, WA\\
\url{www.retrocausal.ai}
}

\begin{document}

\twocolumn[{%
\maketitle
\begin{center}
	\centering
		\includegraphics[width=0.85\linewidth, trim = 0mm 60mm 0mm 0mm, clip]{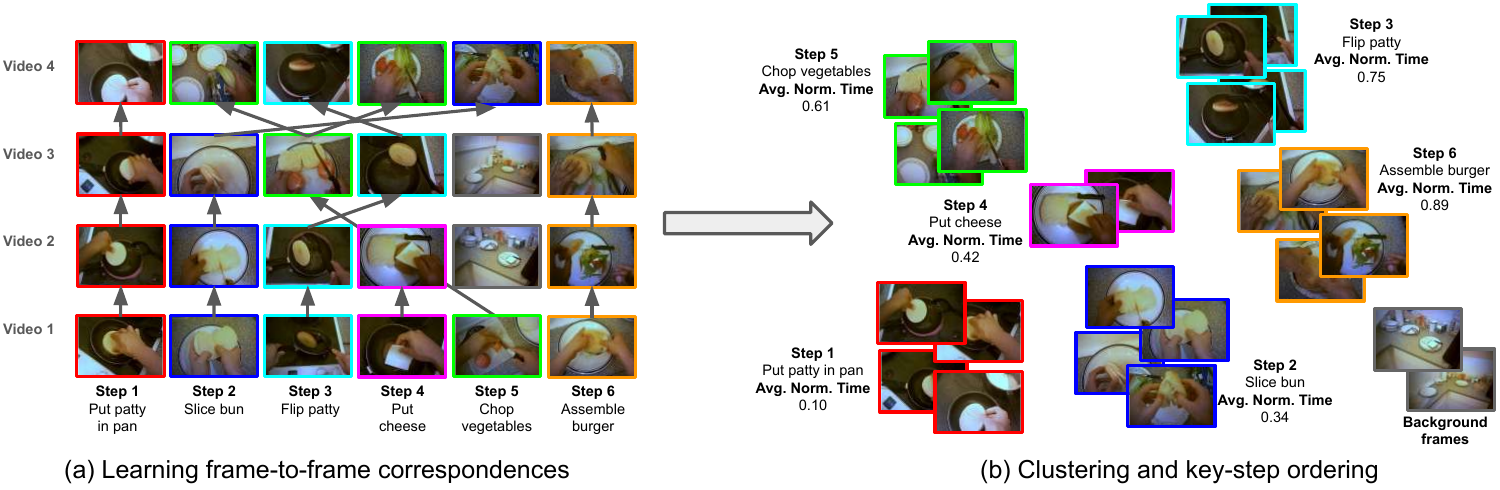}
    \captionof{figure}{Procedure learning methods usually learn video frame representations via temporal video alignment (a). The learned embeddings are then used for extracting key steps and their order (b). In this work, we rely on a regularized Gromov-Wasserstein optimal transport formulation for tackling order variations, background/redundant frames, and repeated actions in (a), yielding state-of-the-art results in (b).}
	\label{fig:teaser}
\end{center}%
}]

\begin{abstract}
We study self-supervised procedure learning, which discovers key steps and their order from a set of unlabeled videos. Previous methods typically learn frame-to-frame correspondences between videos before determining key steps and their order. However, their performance often suffers from order variations, background/redundant frames, and repeated actions. To overcome these challenges, we propose a self-supervised framework, which utilizes a fused Gromov-Wasserstein optimal transport with a structural prior for frame-to-frame mapping. However, optimizing only for the above temporal alignment may lead to degenerate solutions, where all frames are mapped to a small cluster in the embedding space and thus every video is assigned to just one key step. To address that issue, we integrate a contrastive regularization, which maps different frames to various points, avoiding trivial solutions. Finally, extensive experiments on egocentric and third-person benchmarks demonstrate our superior performance over prior works, including OPEL which relies on a classical Kantorovich optimal transport with an optimality prior.
\end{abstract}

\section{Introduction}
\label{sec:introduction}
{\blfootnote{$^{\dagger}$ indicates joint first author.\\ \{ahmed,alishah,umer,fawad,zeeshan,huy\}@retrocausal.ai.}}

Understanding how to perform complex tasks by watching others is a hallmark of human intelligence. An increasingly important goal in computer vision is to replicate this ability in machines, enabling them to learn a procedure from watching video demonstrations. Procedure learning (PL) refers to the task of identifying key steps and their ordering from instructional videos~\cite{alayrac2016unsupervised, kukleva2019unsupervised, zhukov2019cross}. Unlike action recognition and  segmentation, which focus on analyzing single videos and classifying short-term activities~\cite{simonyan2014two, carreira2017quo, piergiovanni2017learning}, procedure learning extracts semantics such as key steps and their order across videos, even when step appearances vary, their order differ, or background/redundant frames occur frequently, as shown in Fig.~\ref{fig:teaser}. These challenges are especially common in real-world datasets.

While earlier methods relied on full or weak supervision~\cite{bojanowski2015weakly, huang2016connectionist} which require substantial human effort and do not scale to large task domains, recent works have focused on self-supervised alignment of instructional videos. CnC~\cite{bansal2022my}, for instance, uses temporal cycle-consistency and contrastive learning to align steps across videos. However, it is sensitive to step order variation and background/redundant frames. More recent works have employed optimal transport (OT) formulations for alignment due to their flexibility. One such method, VAVA~\cite{liu2022learning}, considers alignment as an OT problem to accommodate variations in step ordering and incorporates an optimality prior along with inter-video and intra-video contrastive losses to solve the temporal misalignment problem. However, a key limitation in VAVA~\cite{liu2022learning} is that it is challenging to balance multiple losses and handle repeated actions, as noted by~\cite{donahue2024learning, ali2025joint}. Recently, OPEL~\cite{chowdhuryopel} applies VAVA for self-supervised procedure learning, inheriting these limitations.

In this paper, we propose a self-supervised procedure learning framework built upon a fused Gromov-Wasserstein optimal transport (FGWOT) formulation with a structural prior. Unlike previous OT-based approaches~\cite{liu2022learning,chowdhuryopel} that utilize a traditional Kantorovich OT formulation, our method not only aligns frames based on appearance but also enforces a structural prior (i.e., temporal consistency) across videos. This makes our model better suited for handling step order variations, background/redundant frames, and repeated actions. Moreover, we empirically identify that optimizing only for temporal alignment may lead to degenerate solutions, where all frames collapse to a single cluster in the embedding space. To prevent this issue, we utilize Contrastive Inverse Difference Moment (C-IDM) as a regularization that encourages embedding diversity across frames. Our regularized Gromov-Wasserstein optimal transport (RGWOT) approach uses a unified loss with a purpose-aligned regularization, avoiding the difficulty of balancing multiple losses and conflicting regularizations. Our model outperforms prior works on both egocentric~\cite{bansal2022my} and third-person~\cite{elhamifar2019unsupervised, zhukov2019cross} procedure learning benchmarks..

In summary, our contributions include:
\begin{itemize}
    \item We introduce an optimal transport-based framework for self-supervised procedure learning, which adopts a fused Gromov-Wasserstein optimal transport formulation with a structural prior for frame-to-frame correspondences.
    \item We empirically analyze potential trivial solutions when optimizing only for the above temporal alignment term and incorporate a contrastive regularization term to prevent the collapse to degenerate solutions.
    \item Extensive evaluations on egocentric and third-person datasets show that our approach outperforms prior works, including OPEL which adopts a classical Kantorovich optimal transport formulation with an optimality prior.
\end{itemize}
\section{Related Work}
\label{sec:relatedwork}

\noindent \textbf{Self-Supervised Learning.}
Self-supervised learning uses pretext tasks that generate a supervision signal from the data itself. Early self-supervised approaches focused mainly on images, leveraging spatial cues through tasks such as image colorization~\cite{larsson2016learning,larsson2017colorization}, object counting~\cite{liu2018leveraging, noroozi2017representation}, solving puzzles~\cite{carlucci2019domain, kim2019self, kim2018learning}, predicting rotations~\cite{feng2019self, gidaris2018unsupervised}, and image inpainting~\cite{jenni2020steering}. Recently, self-supervised learning has expanded to video data, which inherently offers spatial and temporal structures. Video-based pretext tasks include forecasting future frames~\cite{ahsan2018discrimnet, diba2019dynamonet, srivastava2015unsupervised, vondrick2016generating, han2019video}, enforcing temporal coherence~\cite{goroshin2015unsupervised, mobahi2009deep, zou2011unsupervised}, predicting the temporal order of frames~\cite{fernando2017self, lee2017unsupervised, misra2016shuffle, xu2019self}, determining the arrow of time~\cite{pickup2014seeing, wei2018learning}, and predicting the pace of actions~\cite{benaim2020speednet, wang2020self, yao2020video}, and frame clustering~\cite{kumar2022unsupervised,tran2024permutation}. In this work, we focus on self-supervised procedure learning for identifying key steps and their order from videos.

\noindent \textbf{Procedure Learning.}
Many procedure learning (PL) methods focus on learning frame-level features that capture task structure~\cite{elhamifar2019unsupervised, kukleva2019unsupervised,elhamifar2020self, vidalmata2021joint,bansal2022my}. For example, Kukleva et al.~\cite{kukleva2019unsupervised} learn frame representation by using timestamps, while VidalMata et al.~\cite{vidalmata2021joint} focus on predicting future frames. Elhamifar et al.~\cite{elhamifar2020self} utilize attention mechanism on individual frames for better feature learning. Bansal et al.~\cite{bansal2022my} use temporal correspondences across videos to create signals for robust frame-level embeddings. Recently, Bansal et al.~\cite{bansal2024united} use a task-level graph representation to cluster frames that are semantically similar and temporally close. Beyond purely visual approaches, procedure learning has also been explored in multi-modal settings, combining narrated text and videos~\cite{damen2014you, yu2014instructional, malmaud2015s,alayrac2016unsupervised, doughty2020action, fried2020learning}, as well as modalities like optical flow, depth, and gaze information~\cite{shah2023steps}. These multi-modal approaches generally assume reliable alignment between modalities~\cite{alayrac2016unsupervised, malmaud2015s, yu2014instructional}, an assumption that frequently fails due to asynchrony among data streams~\cite{elhamifar2019unsupervised,elhamifar2020self}. Also, reliance on imperfect automatic speech recognition requires manual correction, and multi-modal integration adds memory/computation. A recent PL work~\cite{chowdhuryopel} focuses purely on visual data, bypassing the inaccuracies associated with multi-modal alignment. Here, we propose an advanced optimal transport formulation, outperforming~\cite{chowdhuryopel} on visual procedure learning.

\noindent \textbf{Video Alignment.}
TCC~\cite{dwibedi2019tcc} enforces cycle consistency, while GTCC~\cite{donahue2024learning} extends it with multi-cycle consistency to handle repeated actions. However, they focus on local alignment and do not capture the global temporal structure. For global alignment, LAV~\cite{haresh2021learning} applies dynamic time warping~\cite{cuturi2017soft} but assumes monotonic ordering without background/redundant frames. VAVA~\cite{liu2022learning} addresses non-monotonic sequences and background/redundant frames using a classical Kantorovich optimal transport~\cite{cuturi2013sinkhorn} but suffers from balancing multiple losses and repeated actions. Recently, VAOT~\cite{ali2025joint} develops a fused Gromov-Wasserstein optimal transport which can handle order variations, background/redundant frames, and repeated actions. We empirically show a possible degenerate solution for VAOT~\cite{ali2025joint} and propose a regularized optimal transport to address that.

\noindent \textbf{Learning Key-Step Ordering.}
People often perform the same task using different orders of key-steps, but most prior works fail to model this variation. They either assume a strict step ordering~\cite{elhamifar2019unsupervised,kukleva2019unsupervised,vidalmata2021joint} or do not predict the order at all~\cite{elhamifar2020self,shen2021learning}. To better capture how tasks are actually completed, it is important to account for these different key-step sequences. Our method addresses this by creating a specific sequence of key-steps for each video, allowing the model to infer the particular order required to complete the task.

\noindent \textbf{Temporal Contrastive Learning.}
Inspired by the success of contrastive learning for self-supervised image representation learning~\cite{chen2020simple,he2020momentum,caron2020unsupervised}, temporal contrastive learning has been proposed for video representation learning~\cite{qian2021spatiotemporal,basu2022unsupervised,dave2022tclr}. For example, TCLR~\cite{dave2022tclr} applies inter- and intra-instance temporal contrastive losses to capture temporal variations between and within videos. Unlike these methods which use contrastive learning as their main pretext task, we use video alignment as our main pretext task, and contrastive learning as our regularization to avoid trivial solutions.
\section{Our Approach}
\label{sec:method}

\begin{figure*}[t]
	\centering
		\includegraphics[width=0.85\linewidth, trim = 0mm 55mm 30mm 0mm, clip]{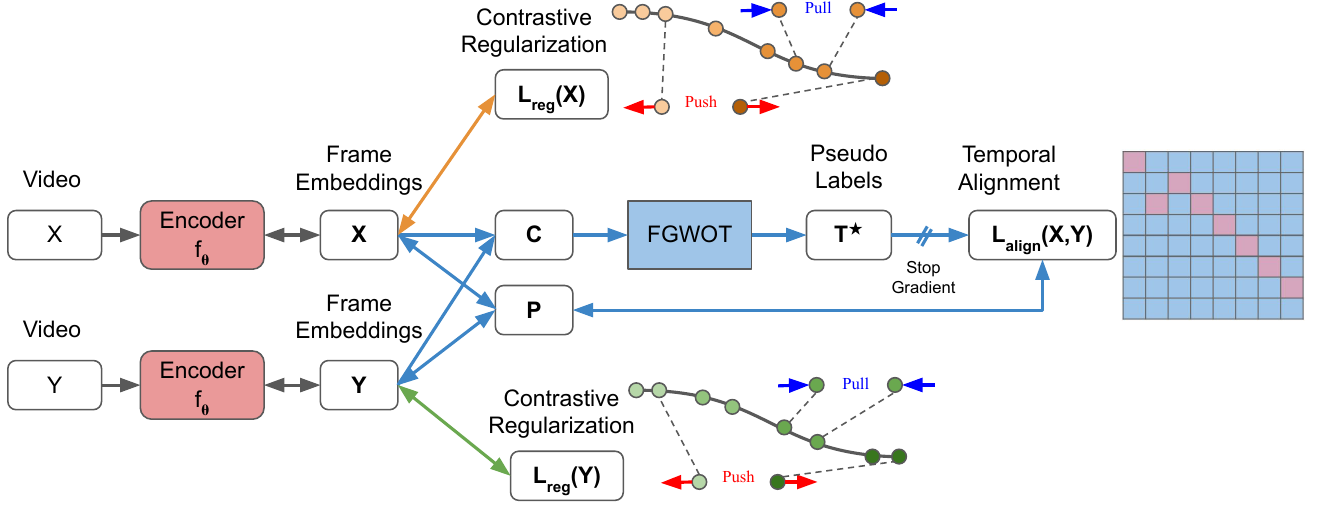}
	\caption{Our approach incorporates a fused Gromov-Wasserstein optimal transport formulation with a structural prior for establishing frame-to-frame correspondences between videos with a contrastive regularization for avoiding degenerate solutions. Forward/backward arrows denote computation/gradient flows. Blue and orange/green represent temporal alignment and contrastive regularization.}
	\label{fig:method}
\end{figure*}

Our approach includes finding frame-to-frame correspondences between videos in Sec.~\ref{sec:alignment} and performing clustering and key-step ordering in Sec.~\ref{sec:clustering}.

\noindent \textbf{Notations.} Let us represent two input videos of $N$ and $M$ frames as $X = \{x_1, \dots, x_N\}$ and $Y = \{y_1, \dots, y_M\}$  respectively. The frame-level embeddings of $X$ and $Y$ are then written as $\mathbf{X} = f_{\bm{\theta}}(X) \in \mathbb{R}^{N \times D}$ and $\mathbf{Y} = f_{\bm{\theta}}(Y) \in \mathbb{R}^{M \times D}$ respectively, where $f_{\bm{\theta}}$ (with learnable parameters $\bm{\theta}$) is the embedding function and $D$ is the length of the embedding vector. $K$ is the number of key steps. In addition, the dot product of $\mathbf{A}, \mathbf{B} \in \mathbb{R}^{n \times m}$ is expressed as $\langle \mathbf{A}, \mathbf{B} \rangle = \sum_{i, j} A_{ij} B_{ij}$, while a vector of ones is written as $\mathbf{1}_n \in \mathbb{R}^n$. Finally, $[n]= \{1, \dots, n\}$ models a discrete set of $n$ elements. We represent the $(m-1)$ dimensional probability simplex as $\Delta_m \subset \mathbb{R}^m$ and the Cartesian product space consisting of $n$ such simplexes as $\Delta_m^n \subset \mathbb{R}^{m \times n}$.

\subsection{Regularized Gromov-Wasserstein Optimal Transport}
\label{sec:alignment}

An overview of our regularized Gromov-Wasserstein optimal transport framework (RGWOT) is shown in Fig.~\ref{fig:method}.

\subsubsection{Fused Gromov-Wasserstein Optimal Transport}

We consider temporal video alignment as an optimal transport problem in the discrete setting. We follow~\cite{ali2025joint} to apply a fused Gromov-Wasserstein optimal transport formulation (FGWOT), which is a \emph{combination} of Kantorovich optimal transport (KOT) and Gromov-Wasserstein optimal transport (GWOT). Below we provide the details of KOT and GWOT. In particular, given a ground cost $\mathbf{C}\in\mathbb{R}^{n\times m}_+$, the KOT objective seeks the minimum-cost coupling $\mathbf{T}^\star$ between histograms $\mathbf{p}\in \Delta_n$ and $\mathbf{q} \in \Delta_m$ as: 
\begin{equation}
\label{eq:ot_prob}
   \argmin_{\mathbf{T}\in \mathcal{T}_{\mathbf{p}, \mathbf{q}}}~~~\mathcal{F}_{\text{KOT}}(\mathbf{C}, \mathbf{T}) = \langle \mathbf{C}, \mathbf{T}\rangle.
\end{equation}
Here, $\mathcal{T}_{\mathbf{p}, \mathbf{q}} = \{\mathbf{T}\in\mathbb{R}_+^{n\times m} \mid \mathbf{T} \mathbf{1}_m = \mathbf{p}, \mathbf{T}^\top \mathbf{1}_n = \mathbf{q} \}$ is known as the transportation polytope and the coupling $\mathbf{T}$ is considered as the \emph{soft assignment} between elements in the supports of $\mathbf{p}$ and $\mathbf{q}$, i.e., discrete sets $[n]$ and $[m]$. For video alignment, $\mathbf{T}$ models the mapping between frames of $X$ and $Y$. Furthermore, given distance matrices $\mathbf{C}^x\in\mathbb{R}^{n\times n}$ and $\mathbf{C}^y\in\mathbb{R}^{m\times m}$ defined over supports $[n]$ and $[m]$ respectively and a cost function $L:\mathbb{R}\times \mathbb{R} \rightarrow \mathbb{R}$ minimizing discrepancies between distance matrix elements, the GWOT objective is written as:
\begin{equation}
\label{eq:gwot_obj}
    \argmin_{\mathbf{T}\in \mathcal{T}_{\mathbf{p}, \mathbf{q}}}~~~\mathcal{F}_{\text{GWOT}}(\mathbf{C}^x, \mathbf{C}^y, \mathbf{T}) = \sum_{\substack{i,k \in [n]\\j,l \in [m]}} L(\mathbf{C}_{ik}^x, \mathbf{C}_{jl}^y) \mathbf{T}_{ij}\mathbf{T}_{kl}.
\end{equation}
For video alignment, the GWOT objective enforces \emph{structural priors} $\mathbf{C}^x$ and $\mathbf{C}^y$ (i.e., temporal consistency) on the transport map, which will be elaborated below. Finally,  given a balancing parameter $\alpha \in [0, 1]$, the FGWOT objective fuses the KOT and GWOT objectives as:
\begin{multline}
\label{eq:fgwot_obj}
   \argmin_{\mathbf{T}\in \mathcal{T}_{\mathbf{p}, \mathbf{q}}}~~~\mathcal{F}_{\text{FGWOT}}(\mathbf{C}, \mathbf{C}^x, \mathbf{C}^y, \mathbf{T}) =  (1-\alpha) \mathcal{F}_{\text{KOT}}(\mathbf{C}, \mathbf{T}) +\\ \alpha\mathcal{F}_{\text{GWOT}}(\mathbf{C}^x, \mathbf{C}^y, \mathbf{T}).
\end{multline}
For video alignment, the KOT objective minimizes visual differences between corresponding frames of $X$ and $Y$, while the GWOT objective imposes structural properties on the resulting mapping (i.e., temporal consistency). We denote $\mathbf{p} = \frac{1}{N}\mathbf{1}_N$ and $\mathbf{q} = \frac{1}{M}\mathbf{1}_M$ as histograms defined over the sets of $N$ frames in $X$ and $M$ frames in $Y$, represented by $[N]$ and $[M]$ respectively. The solution $\mathbf{T}^\star\in\mathbb{R}_+^{N\times M}$ between $[N]$ and $[M]$ models the soft assignment between frames of $X$ and $Y$. Below we will discuss the cost matrices $\{\mathbf{C}, \mathbf{C}^x, \mathbf{C}^y\}$, obtaining the solution $\mathbf{T}^\star$ efficiently and robustly, and self-supervised training.  

\noindent \textbf{Cost Matrices.}
The cost matrix $\mathbf{C}\in\mathbb{R}^{N\times M}_+$ in the KOT component measures visual differences between $X$ and $Y$ and is defined as $\mathbf{C}_{ij} = 1 - \frac{\mathbf{x}_i^\top \mathbf{y}_j}{\|\mathbf{x}_i\|_2\|\mathbf{y}_j\|_2}$, with frame embeddings $\mathbf{x}_i = f_{\bm{\theta}}(x_i)$ and $\mathbf{y}_j = f_{\bm{\theta}}(y_j)$. Also, we follow~\cite{liu2022learning} to augment $\mathbf{C}$ as $\Tilde{\mathbf{C}} = \mathbf{C} + \rho \mathbf{R}$, where $\rho \geq 0$ is a balancing parameter and the \emph{temporal prior} $\mathbf{R}$ (with $\mathbf{R}_{ij} = \left| i / N - j / M \right|$) enforces the coupling $\mathbf{T}$ to have a banded diagonal shape. Moreover, we set $L(a,b) = ab$, and follow~\cite{ali2025joint} to define cost matrices $\mathbf{C}^x\in\mathbb{R}_+^{N\times N}$ and $\mathbf{C}^y\in\mathbb{R}_+^{M\times M}$ in the GWOT component as:
\begin{equation}
\label{eq:cxcy_def}
    \mathbf{C}^x_{ik} =
    \begin{cases}
        \frac{1}{r} & 1 \leq \delta_{ik} \leq Nr \\
        0 & \text{otherwise}
    \end{cases},
    \mathbf{C}^y_{jl} =
    \begin{cases}
        0 & 1 \leq \delta_{jl} \leq Mr \\
        1 & \text{otherwise}
    \end{cases},
\end{equation}
with $\delta_{ik} = |i-k|$, $\delta_{jl} = |j-l|$, and a radius parameter $r \in (0,1]$. The GWOT component imposes temporal consistency over $\mathbf{T}$. Specifically, mapping temporally adjacent frames in $X$ ($\delta_{ik} \leq Nr$) to temporally remote frames in $Y$ ($\delta_{jl} > Mr$) induces a cost ($L(\mathbf{C}^x_{ik},\mathbf{C}^y_{jl}) = \frac{1}{r}$), however assigning temporally neighboring frames in $X$ ($\delta_{ik} \leq Nr$) to temporally nearby frames in $Y$ ($\delta_{jl} \leq Mr$) or assigning temporally remote frames in $X$ ($\delta_{ik} > Nr$) to temporally distant frames in $Y$ ($\delta_{jl} > Mr$) induces no cost ($L(\mathbf{C}^x_{ik},\mathbf{C}^y_{jl}) = 0$). The GWOT component is able to tackle order variations and repeated actions~\cite{xu2024temporally}.

\noindent \textbf{Efficient and Robust Solution.}
Following~\cite{peyre2016gromov}, we can compute the GWOT component efficiently as $\mathcal{F}_{\text{GWOT}}(\mathbf{C}^x, \mathbf{C}^y, \mathbf{T}) = \langle \mathbf{C}^x \mathbf{T} \mathbf{C}^y , \mathbf{T} \rangle$. By adding an entropy regularization term $-\epsilon H(\mathbf{T})$, with $H(\mathbf{T}) = -\sum_{i,j} T_{ij}\log T_{ij}$ and $\epsilon > 0$, to the FGWOT formulation in Eq.~\ref{eq:fgwot_obj}, the solution $\mathbf{T}^\star$ can be derived efficiently, following~\cite{peyre2016gromov}. Our solver often needs less than $25$ iterations to converge, and thanks to the sparse $\mathbf{C}^x$ and $\mathbf{C}^y$, each iteration has $O(NM)$ time complexity. In addition, to enhance robustness, we follow~\cite{liu2022learning} to append a \emph{virtual frame} to $X$ and $Y$ for tackling background/redundant frames. In particular, we include an additional row and column to $\mathbf{T}$ and expand other variables accordingly. If the matching probability of $x_i$ ($i \leq N$) with every $y_j$ ($j \leq M$) is below a threshold parameter $\zeta$, we assign $x_i$ to the virtual frame $y_{M+1}$, and vice versa. Note that virtual frames and their corresponding frames are not included in computing the losses.

\noindent \textbf{Self-Supervised Training.}
We train the frame encoder $f_{\bm{\theta}}$ via the cross-entropy loss between normalized similarities $\mathbf{P}$ (computed based on frame embeddings $\mathbf{X}$ and $\mathbf{Y}$) and pseudo-labels $\mathbf{T}^\star$ (derived from the FGWOT formulation in Eq.~\ref{eq:fgwot_obj}). We use $\mathbf{P}_{ij} = \frac{\exp(\mathbf{X} \mathbf{Y}^\top / \tau)_{ij}}{\sum_l\exp(\mathbf{X} \mathbf{Y}^\top / \tau)_{il}}$, with a temperature parameter $\tau > 0$. Our self-supervised loss is written as:
\begin{equation}
\label{eq:lxy}
    \mathcal{L}_{align}(\mathbf{X},\mathbf{Y}) = -\sum_{i=1}^N\sum_{j=1}^M \mathbf{T}^\star_{ij} \log \mathbf{P}_{ij}.
\end{equation}

\subsubsection{Degenerate Solution}
\begin{figure}[b]
	\centering
		\includegraphics[width=0.95\linewidth, trim = 0mm 100mm 140mm 0mm, clip]{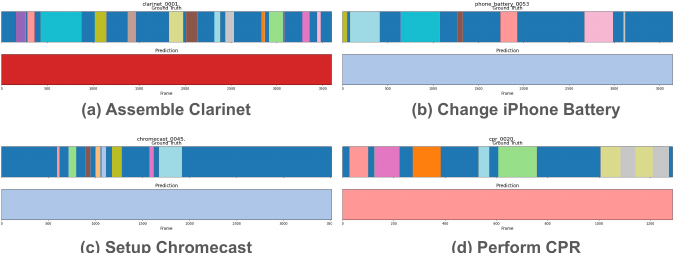}
	\caption{Degenerate solutions by~\cite{ali2025joint} across four different subtasks of ProceL~\cite{elhamifar2019unsupervised}. Ground truth and results by~\cite{ali2025joint} are shown at top and bottom rows respectively.}
	\label{fig:degenerate}
\end{figure}

Although the FGWOT formulation has demonstrated great performance on several video alignment benchmarks, including Pouring, Penn Action, and IKEA ASM, in~\cite{ali2025joint}, optimizing exclusively for the FGWOT objective may potentially lead to trivial solutions. More specifically, since the cost matrix $\mathbf{C}$ minimizes visual differences between corresponding frames in $X$ and $Y$, and both the temporal prior $\mathbf{R}$ and structural priors $\mathbf{C}^x$ and $\mathbf{C}^y$ minimizes temporal differences between corresponding frames in $X$ and $Y$ (e.g., temporally nearby frames in $X$ should be paired with temporally adjacent frames in $Y$ and vice versa), there is no mechanism to prevent the optimization from collapsing to degenerate solutions, where all frames are mapped to a small cluster in the embedding space, and hence an entire video is assigned to a single key step. We consistently observe this issue on ProceL and illustrate some qualitative results in Fig.~\ref{fig:degenerate}. From the results, every video is associated with only one key step, whereas the ground truth contains many. We note that due to different dataset characteristics, this problem seldom appears on EgoproceL and CrossTask.

\subsubsection{Contrastive Regularization}
To avoid trivial solutions, we incorporate Contrastive Inverse Difference Moment (C-IDM)~\cite{haresh2021learning} as a regularization, which is applied separately on frame embeddings $\mathbf{X}$ and $\mathbf{Y}$, yielding our regularized Gromov-Wasserstein optimal transport framework (RGWOT), as illustrated in Fig.~\ref{fig:method}. Since $\mathcal{L}_{reg}(\mathbf{X})$ and $\mathcal{L}_{reg}(\mathbf{Y})$ are similar, below we describe $\mathcal{L}_{reg}(\mathbf{X})$ only, which is written as:
\begin{equation}
\label{eq:idm_contrastive}
\begin{split}
    \mathcal{L}_{reg}(\mathbf{X}) = \sum_{i=1}^{N} \sum_{j=1}^{N} (1-\eta(i,j)) \omega(i,j)  max(0,\lambda - d_{\mathbf{X}}(i,j)) \\
    + \eta(i,j) \frac{d_{\mathbf{X}}(i,j)}{\omega(i,j)}.
\end{split}
\end{equation}
Here, the neighboring indicator $\eta(i,j) = 1$ if $|i-j| \leq \sigma$ and 0 otherwise, $\sigma$ is a window size, the temporal weight $\omega(i,j) = (i-j)^2 + 1$, $\lambda$ is a margin parameter, and $d_{\mathbf{X}}(i,j) = \|\mathbf{x}_i-\mathbf{x}_j\|_2$. $\mathcal{L}_{reg}(\mathbf{X})$ encourages temporally nearby frames in $X$ to be mapped to adjacent points in the embedding space, while penalizing temporally distant frames in $X$ when their distance in the embedding space is smaller $\lambda$. As a result, $\mathcal{L}_{reg}(\mathbf{X})$ and $\mathcal{L}_{reg}(\mathbf{Y})$ enforce diverse frame embeddings $\mathbf{X}$ and $\mathbf{Y}$ respectively, effectively preventing the collapse to degenerate solutions.

Finally, our training objective is written as:
\begin{equation}
\label{eq:final}
    \mathcal{L} = \mathcal{L}_{align}(\mathbf{X},\mathbf{Y}) + \beta \left( \mathcal{L}_{reg}(\mathbf{X}) + \mathcal{L}_{reg}(\mathbf{Y}) \right),
\end{equation}
where $\beta$ is the weight for the contrastive regularization.

\subsection{Clustering and Key-Step Ordering}
\label{sec:clustering}

Given frame embeddings learned via the above RGWOT approach, our next steps include localizing key steps and determining their order for procedure learning. We model key step localization as multi-label graph-cut segmentation~\cite{greig1989exact}, following~\cite{bansal2022my}. We first build a graph which includes $K$ terminal nodes representing $K$ key steps and non-terminal nodes representing frame embeddings, as well as t-links connecting non-terminal nodes to terminal nodes and n-links connecting two non-terminal nodes. We employ $\alpha$-Expansion~\cite{boykov2002fast} to obtain the optimal cut, yielding frame assignments to $K$ key steps. Next, for each video, the order of key steps are determined by sorting the average normalized time of their assigned frames, following~\cite{bansal2022my}. Finally, we rank the key-step orders based on their frequency appearing in the videos and output the top ranked key-step order.
\section{Experiments}
\label{sec:experiments}

\noindent \textbf{Datasets.}
We test our RGWOT approach on three datasets: one first-person dataset, EgoProceL~\cite{bansal2022my}, and two third-person datasets, ProceL~\cite{elhamifar2019unsupervised} and CrossTask~\cite{zhukov2019cross}. EgoProceL contains 62 hours of egocentric videos capturing 16 diverse tasks. ProceL contains 720 videos of 12 tasks with visually similar key-steps that span a total of 47 hours. CrossTask consists of 2763 videos totaling 213 hours for 18 tasks. Following previous works~\cite{bansal2022my,bansal2024united,chowdhuryopel}, the same training and validation splits were used for all methods.

\noindent \textbf{Competing Methods.} 
We compare our RGWOT method with previous self-supervised procedure learning methods, as well as with random and uniform distributions. Competing methods include CNC~\cite{bansal2022my}, GPL~\cite{bansal2024united} and OPEL~\cite{chowdhuryopel}. OPEL, in particular, uses a Kantorovich optimal transport formulation, making it our closest competitor.

\noindent \textbf{Evaluation Metrics.} We evaluate our RGWOT approach for procedure learning on the validation set of all datasets in accordance with prior works~\cite{bansal2022my, bansal2024united, chowdhuryopel}. For the egocentric dataset (EgoProceL~\cite{bansal2022my}), we report F1-score and Intersection over Union (IoU). Furthermore, to evaluate our RGWOT approach on third-person datasets (ProceL~\cite{elhamifar2019unsupervised} and CrossTask~\cite{zhukov2019cross}), we report precision, recall, and F1-score. Framewise scores are computed separately for each keystep, and the final score is reported as the average across all keysteps. We apply the Hungarian matching algorithm~\cite{kuhn1955hungarian} to align the predicted labels with the ground truth, similarly to prior works ~\cite{ elhamifar2019unsupervised, elhamifar2020self, shen2021learning, bansal2022my, bansal2024united, chowdhuryopel}.

\noindent \textbf{Implementation Details.}
We adopt ResNet-50 ~\cite{he2016deep} as the backbone network for feature extraction. We trained the encoder network using a pair of videos, randomly sampling frames from each during training as done by~\cite{dwibedi2019tcc}. The network is optimized with the proposed RGWOT loss. Features are taken from the Conv4c layer, and a sequence of $c$ context frame features is stacked along the temporal axis. To encode temporal information, the stacked features are fed into two 3D convolutional layers, followed by a 3D global max pooling layer. The resulting representation is then passed through two fully connected layers and finally projected into a 128-dimensional embedding space using a linear layer. We follow the task-specific protocols from~\cite{elhamifar2020self,bansal2022my,chowdhuryopel} in all experiments.

\subsection{State-of-the-Art Comparison Results}

\noindent \textbf{Comparisons on Egocentric Dataset.}
Tab.~\ref{tab:egoprocel_comparison} provides comparative evaluation of our RGWOT approach against previous methods on the large-scale egocentric (first-person) benchmark of EgoProceL~\cite{bansal2022my}. In particular, it is a recent dataset tailored for egocentric procedure learning, serving as a strong benchmark for evaluating new methods on egocentric videos. Remarkably, RGWOT consistently surpasses existing approaches across all sub-datasets of EgoProceL~\cite{bansal2022my}. Notably, RGWOT achieves an average improvement of \textbf{15.1\%} in F1 score and \textbf{15.3\%} in IoU compared to previous methods.  These results demonstrate the superiority of our RGWOT formulation with a structural prior over OPEL~\cite{chowdhuryopel}'s traditional Kantorovich optimal transport formulation with an optimality prior.
\begin{table*}[t]
    \centering
    \small
    \caption{Comparisons on egocentric dataset (i.e., EgoProceL~\cite{bansal2022my}). Best results are in \textbf{bold}, while second best are \underline{underlined}.}
    \setlength{\tabcolsep}{1.5pt}
    \begin{tabular}{@{}lcccccccccccccc@{}}\toprule
        & \multicolumn{12}{c}{EgoProceL~\cite{bansal2022my}} \\
        \cmidrule(lr){2-13}
        & \multicolumn{2}{c}{CMU-MMAC~\cite{de2009guide}} & \multicolumn{2}{c}{EGTEA-GAZE+~\cite{li2018eye}} & \multicolumn{2}{c}{MECCANO~\cite{ragusa2021meccano}} & \multicolumn{2}{c}{EPIC-Tents~\cite{jang2019epic}} & \multicolumn{2}{c}{PC Assembly} & \multicolumn{2}{c}{PC Disassembly} \\
        \cmidrule(lr){2-3}\cmidrule(lr){4-5}\cmidrule(lr){6-7}\cmidrule(lr){8-9}\cmidrule(lr){10-11}\cmidrule(lr){12-13}
         & F1 & IoU & F1 & IoU & F1 & IoU & F1 & IoU & F1 & IoU & F1 & IoU \\
        \midrule
        Random & 15.7 & 5.9 & 15.3 & 4.6 & 13.4 & 5.3 & 14.1 & 6.5 & 15.1 & 7.2 & 15.3 & 7.1 \\
        Uniform & 18.4 & 6.1 & 20.1 & 6.6 & 16.2 & 6.7 & 16.2 & 7.9 & 17.4 & 8.9 & 18.1 & 9.1 \\
        CnC~\cite{bansal2022my} & 22.7 & 11.1 & 21.7 & 9.5 & 18.1 & 7.8 & 17.2 & 8.3 & 25.1 & 12.8 & 27.0 & 14.8 \\
        GPL-2D~\cite{bansal2024united} & 21.8 & 11.7 & 23.6 & 14.3 & 18.0 & 8.4 & 17.4 & 8.5 & 24.0 & 12.6 & 27.4 & 15.9 \\
        UG-I3D~\cite{bansal2024united} & 28.4 & 15.6 & 25.3 & 14.7 & 18.3 & 8.0 & 16.8 & 8.2 & 22.0 & 11.7 & 24.2 & 13.8 \\
        GPL-w BG~\cite{bansal2024united} & 30.2 & 16.7 & 23.6 & 14.9 & 20.6 & 9.8 & 18.3 & 8.5 & 27.6 & 14.4 & 26.9 & 15.0 \\
        GPL-w/o BG~\cite{bansal2024united} & 31.7 & 17.9 & 27.1 & 16.0 & 20.7 & 10.0 & 19.8 & 9.1 & 27.5 & 15.2 & 26.7 & 15.2 \\
        OPEL~\cite{chowdhuryopel} & 36.5 & 18.8 & 29.5 & 13.2 & 39.2 & 20.2 & 20.7 & 10.6 & 33.7 & 17.9 & 32.2 & 16.9 \\
        VAOT~\cite{ali2025joint} & \underline{50.3} & \underline{34.9} & \underline{35.1} & \underline{20.4} & \underline{57.3} & \underline{41.2} & \underline{33.5} & \underline{20.4} & \underline{38.1} & \underline{23.6} & \underline{44.4} & \underline{28.9} \\
        RGWOT (Ours) & \textbf{54.4} & \textbf{38.6} & \textbf{37.4} & \textbf{22.9} & \textbf{59.5} & \textbf{42.7} & \textbf{39.7} & \textbf{24.9} & \textbf{43.6} & \textbf{28.0} & \textbf{45.9} & \textbf{30.1} \\
        \bottomrule
    \end{tabular}
    \label{tab:egoprocel_comparison}
\end{table*}

\noindent \textbf{Comparisons on Third-Person Datasets.}
In Tab.~\ref{tab:third_person_comparison}, we evaluate the performance of our RGWOT approach against previous self-supervised procedure learning models on two third-person datasets, i.e., ProceL~\cite{elhamifar2019unsupervised} and CrossTask~\cite{zhukov2019cross}. It is evident that RGWOT achieves the best overall results across both datasets, outperforming all existing models. Notably, RGWOT demonstrates a significant improvement over OPEL~\cite{chowdhuryopel}, the previous best performing approach. Our approach surpasses OPEL~\cite{chowdhuryopel} by \textbf{9.4\%} on ProceL~\cite{elhamifar2019unsupervised} and by \textbf{5.4\%} on CrossTask\cite{zhukov2019cross} on the F1 scores.

\begin{table}[t]
    \centering
    \small
    \caption{Comparisons on third-person datasets (i.e., ProceL~\cite{elhamifar2019unsupervised} and CrossTask~\cite{zhukov2019cross}). Best results are in \textbf{bold}, while second best are \underline{underlined}. `X' denotes degenerate results.}
    \setlength{\tabcolsep}{4pt}
    \begin{tabular}{@{}lrrrrrr@{}}
    \toprule
    & \multicolumn{3}{c}{ProceL~\cite{elhamifar2019unsupervised}} & \multicolumn{3}{c}{CrossTask~\cite{zhukov2019cross}} \\
    \cmidrule(lr){2-4} \cmidrule(l){5-7}
    & P & R & F & P & R & F \\
    \midrule
    Uniform & 12.4 & 9.4 & 10.3 & 8.7 & 9.8 & 9.0 \\
    Alayrac \etal~\cite{alayrac2016unsupervised} & 12.3 & 3.7 & 5.5 & 6.8 & 3.4 & 4.5 \\
    Kukleva \etal~\cite{kukleva2019unsupervised} & 11.7 & 30.2 & 16.4 & 9.8 & 35.9 & 15.3 \\
    Elhamifar \etal~\cite{elhamifar2020self} & 9.5 & 26.7 & 14.0 & 10.1 & \textbf{41.6} & 16.3 \\
    Fried \etal~\cite{fried2020learning} & -- & -- & -- & -- & 28.8 & -- \\
    Shen \etal~\cite{shen2021learning} & 16.5 & 31.8 & 21.1 & 15.2 & 35.5 & 21.0 \\
    CnC~\cite{bansal2022my} & 20.7 & 22.6 & 21.6 & 22.8 & 22.5 & 22.6 \\
    GPL-2D~\cite{si2020adversarial} & 21.7 & 23.8 & 22.7 & 24.1 & 23.6 & 23.8 \\
    UG-I3D~\cite{si2020adversarial} & 21.3 & 23.0 & 22.1 & 23.4 & 23.0 & 23.2 \\
    GPL~\cite{si2020adversarial} & 22.4 & 24.5 & 23.4 & 24.9 & 24.1 & 24.5 \\
    STEPs~\cite{shah2023steps} & 23.5 & 26.7 & 24.9 & 26.2 & 25.8 & 25.9 \\
    OPEL~\cite{chowdhuryopel} & \underline{33.6} & \underline{36.3} & \underline{34.9} & 35.6 & 34.8 & 35.1 \\
    VAOT~\cite{ali2025joint} & X & X & X & \underline{38.2} & 38.9 & \underline{38.6} \\
    RGWOT (Ours) & \textbf{42.2} & \textbf{46.7} & \textbf{44.3} & \textbf{40.4} & \underline{40.7} & \textbf{40.4} \\
    \bottomrule
    \end{tabular}
    \label{tab:third_person_comparison}
\end{table}

\begin{table*}[t]
    \centering
    \small
    \caption{Comparisons against multimodal method on EgoProceL~\cite{bansal2022my}. Best results are in \textbf{bold}, while second best are \underline{underlined}.}
    \setlength{\tabcolsep}{1.5pt}
    \begin{tabular}{@{}lcccccccccccccc@{}}\toprule
        & \multicolumn{2}{c}{CMU-MMAC~\cite{de2009guide}} & \multicolumn{2}{c}{EGTEA-GAZE+~\cite{li2018eye}} & \multicolumn{2}{c}{MECCANO~\cite{ragusa2021meccano}} & \multicolumn{2}{c}{EPIC-Tents~\cite{jang2019epic}} & \multicolumn{2}{c}{ProceL~\cite{elhamifar2019unsupervised}} & \multicolumn{2}{c}{CrossTask~\cite{zhukov2019cross}} \\
        \cmidrule(lr){2-3}\cmidrule(lr){4-5}\cmidrule(lr){6-7}\cmidrule(lr){8-9}\cmidrule(lr){10-11}\cmidrule(lr){12-13}
         & F1 & IoU & F1 & IoU & F1 & IoU & F1 & IoU & F1 & IoU & F1 & IoU \\
        \midrule
        STEPs~\cite{shah2023steps} & \underline{28.3} & \underline{11.4} & \underline{30.8} & \underline{12.4} & \underline{36.4} & \underline{18.0} & \textbf{42.2} & \underline{21.4} & \underline{24.9} & \underline{15.4} & \underline{25.9} & \underline{14.6} \\
        RGWOT (Ours) & \textbf{54.4} & \textbf{38.6} & \textbf{37.4} & \textbf{22.9} & \textbf{59.5} & \textbf{42.7} & \underline{39.7} & \textbf{24.9} & \textbf{44.3} & \textbf{29.4} & \textbf{40.4} & \textbf{26.3} \\
        \bottomrule
    \end{tabular}
    \label{tab:multimodal_comparison}
\end{table*}

\noindent \textbf{Comparisons with Multimodal Method.}
Tab.~\ref{tab:multimodal_comparison} highlights the performance of our RGWOT approach compared to STEPs~\cite{shah2023steps}, a multimodal approach to unsupervised procedure learning. STEPs~\cite{shah2023steps} utilizes depth and gaze data along with RGB while our RGWOT approach uses only RGB data. Our RGWOT approach outperforms STEPS~\cite{shah2023steps} on most datasets. RGWOT only has lower F1 score on EPIC-Tents~\cite{jang2019epic}, while still outperforming it on IoU. It should also be noted that our approach outperforms models that utilize narrations along with video data, namely Alayrac et al.~\cite{alayrac2016unsupervised} and Shen et al.~\cite{shen2021learning}, as observed in Tab.~\ref{tab:third_person_comparison}.

\noindent \textbf{Comparisons with Action Segmentation Methods.}
The difference between procedure learning (PL) and action segmentation (AS) is that PL is applied to a set of videos performing the same task, where it assigns frames to $K$ key steps and determines their sequence. This allows PL to uncover common procedural structures and identify key steps that repeat consistently across different videos. In contrast, AS operates on a single video, segmenting it into distinct actions without leveraging information from other videos.
\begin{table}[t]
    \centering
    \small
    \caption{Comparisons against action segmentation methods on ProceL~\cite{elhamifar2019unsupervised} and CrossTask~\cite{zhukov2019cross}. Best results are in \textbf{bold}, while second best are \underline{underlined}.}
    \setlength{\tabcolsep}{3.4pt}
    \begin{tabular}{l ccc ccc}
        \toprule
        
        \multirow{3}{*}{\shortstack{Action Segmentation\\Method}} & \multicolumn{3}{c}{ProceL~\cite{elhamifar2019unsupervised}} & \multicolumn{3}{c}{CrossTask~\cite{zhukov2019cross}} \\
        \cmidrule(lr){2-4} \cmidrule(lr){5-7}
        & P & R & F1 & P & R & F1 \\
        \midrule
        JointSeqFL~\cite{elhamifar2019unsupervised} & - & - & 29.8 & - & - & - \\
        Elhamifar et al.~\cite{elhamifar2020self} & 9.5 & 26.7 & 14.0 & 10.1 & \underline{41.6} & 16.3 \\
        Fried et al.~\cite{fried2020learning} & - & - & - & - & 28.8 & - \\
        Shen et al.~\cite{shen2021learning} & 16.5 & 31.8 & 21.1 & 15.2 & 35.5 & 21.0 \\
        Dvornik et al.~\cite{dvornik2022flow} & - & - & - & - & - & 25.3 \\        StepFormer~\cite{dvornik2023stepformer} & 18.3 & 28.1 & 21.9 & 22.1 & \textbf{42.0} & 28.3 \\
        OPEL~\cite{chowdhuryopel} & \underline{33.6} & \underline{36.3} & \underline{34.9} & \underline{35.6} & 34.8 & \underline{35.1} \\
        RGWOT (Ours) & \textbf{42.2} & \textbf{46.7} & \textbf{44.3} & \textbf{40.4} & 40.7 & \textbf{40.4} \\
        \bottomrule
    \end{tabular}
    \label{tab:action_seg_comparison}
\end{table}

Tab.~\ref{tab:action_seg_comparison} presents a comparison of our RGWOT approach against several state-of-the-art unsupervised action segmentation methods, as well as OPEL~\cite{chowdhuryopel}, on the ProceL~\cite{elhamifar2019unsupervised} and CrossTask~\cite{zhukov2019cross} datasets. On the ProceL~\cite{elhamifar2019unsupervised} dataset, RGWOT achieves a substantial improvement with the highest precision, recall, and F1 score, significantly surpassing previous methods. On the CrossTask~\cite{zhukov2019cross} dataset, although StepFormer~\cite{dvornik2023stepformer} achieves the highest recall, its F1 score remains relatively low at \textbf{28.3\%}, indicating a poor balance between precision and recall. In contrast, RGWOT achieves the highest F1 score of \textbf{40.4\%} with a well-balanced precision (\textbf{40.4\%}) and recall (\textbf{40.7\%}), which is critical since F1 provides a more reliable measure of overall segmentation quality. These results demonstrate that RGWOT is robust and effective, consistently outperforming other approaches across both datasets.

\noindent \textbf{Qualitative Results.}
\begin{figure*}[t]
	\centering
		\includegraphics[width=0.95\linewidth]{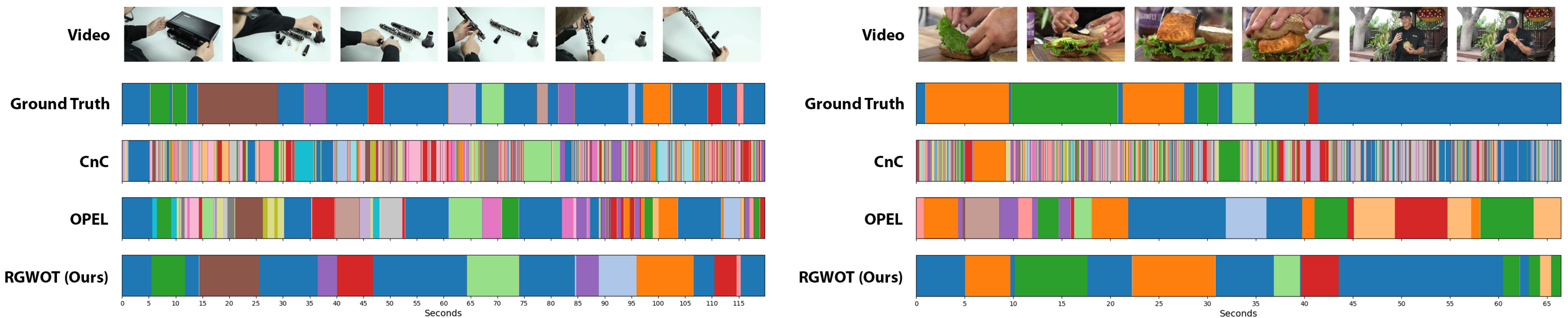}
	\caption{Qualitative results on ProceL~\cite{elhamifar2019unsupervised}. Colored segments represent predicted actions with a particular color denoting the same action across all models.}
	\label{fig:procel_qualitative_results}
\end{figure*}
Fig.~\ref{fig:procel_qualitative_results} illustrates example results of unsupervised procedure learning on two ProceL~\cite{elhamifar2019unsupervised} videos. The CnC~\cite{bansal2022my} model tends to over-segment, and has generally worse alignment for the clusters it predicts correctly. OPEL~\cite{chowdhuryopel} fares better but it also over-segments and misaligns actions. In contrast, our RGWOT model aligns tasks significantly closer to the ground truth and avoids over-segmentation. This visualizes the superior procedure learning capabilities of RGWOT over previous works.

\subsection{Ablation Study Results}

\noindent \textbf{Effects of Model Components.}
Tab.~\ref{tab:ablation_components} presents an ablation study where we systematically remove one model component at a time to assess its contribution to the overall model performance. The baseline configuration includes all four components: contrastive regularization, temporal prior, structural prior, and virtual frame, achieving the highest results. Removing the virtual frame component leads to a substantial performance drop, indicating its critical role. The absence of the temporal and structural priors also results in moderate declines, underscoring their importance. Excluding contrastive regularization causes a smaller degradation, suggesting it is less important on EgoProceL~\cite{bansal2022my}. Nevertheless, contrastive regularization is critical in preventing degenerate solutions on ProceL~\cite{elhamifar2019unsupervised}, as shown in Fig.~\ref{fig:degenerate}. Overall, incorporating all components yields the most robust performance, thereby justifying their inclusion in the complete architecture.
\begin{table*}[t]
    \centering
    \small
    \caption{Effects of model components on performance across MECCANO~\cite{ragusa2021meccano} and PC Disassembly. Best results are in \textbf{bold}, while second best are \underline{underlined}.}
    \setlength{\tabcolsep}{6.3pt}
    \begin{tabular}{cccc cc cc}
        \toprule
        \multirow{2.5}{*}{Contrastive Regularization} & 
        \multirow{2.5}{*}{Temporal Prior} & 
        \multirow{2.5}{*}{Structural Prior} & 
        \multirow{2.5}{*}{Virtual Frame} & 
        \multicolumn{2}{c}{MECCANO~\cite{ragusa2021meccano}} & 
        \multicolumn{2}{c}{PC Disassembly} \\
        \cmidrule(lr){5-6} \cmidrule(lr){7-8}
        & & & & F1 & IoU & F1 & IoU \\
        \midrule
        \checkmark & \checkmark & \checkmark &            & 38.5 & 24.0 & 34.9 & 21.2 \\
        \checkmark & \checkmark &             & \checkmark & 47.1 & 31.4 & 42.9 & 27.1 \\
        \checkmark &            & \checkmark & \checkmark & 51.8 & 35.5 & 43.9 & 27.9 \\
                   & \checkmark & \checkmark & \checkmark & \underline{57.3} & \underline{41.2} & \underline{44.4} & \underline{28.9} \\
        \checkmark & \checkmark & \checkmark & \checkmark & \textbf{59.5} & \textbf{42.7} & \textbf{45.9} & \textbf{30.1} \\
        \bottomrule
    \end{tabular}
    \label{tab:ablation_components}
\end{table*}


\noindent \textbf{Effects of Clustering Methods.}
To evaluate the impact of the clustering method, we replaced our proposed method with K-Means, Subset Selection (SS), and a Random approach where labels are assigned by sampling uniformly over $K$ key steps. As shown in Tab.~\ref{tab:ablation_clustering}, RGWOT achieves the best results across all datasets, highlighting the critical role of our clustering method in the overall framework.
\begin{table*}[t]
    \centering   
    \small
    \caption{Effects of clustering methods on performance across various datasets. Best results are in \textbf{bold}, while second best are \underline{underlined}.}
    \setlength{\tabcolsep}{4pt}
    \begin{tabular}{@{}lcccccccc@{}}\toprule
        & \multicolumn{2}{c}{CMU-MMAC~\cite{de2009guide}} & \multicolumn{2}{c}{EGTEA-GAZE+~\cite{li2018eye}} & \multicolumn{2}{c}{MECCANO~\cite{ragusa2021meccano}} & \multicolumn{2}{c}{EPIC-Tents~\cite{jang2019epic}} \\
        \cmidrule(lr){2-3}\cmidrule(lr){4-5}\cmidrule(lr){6-7}\cmidrule(lr){8-9}
         & F1 & IoU & F1 & IoU & F1 & IoU & F1 & IoU \\
        \midrule
        Random &  16.5 & 9.0 & 15.5 & 8.4 & 14.1 & 7.6 & 14.5 & 7.8 \\
        OT + K-means &  32.8 & 20.0 & 30.3 & 18.2 & 29.5 & 17.4 & 23.8 & 13.5 \\
        OT + SS &  \underline{41.3} & \underline{25.3} & \underline{34.1} & \underline{20.7} & \underline{36.3} & \underline{22.2} & \underline{26.4} & \underline{15.2} \\
        RGWOT & \textbf{54.4} & \textbf{38.6} & \textbf{37.4} & \textbf{22.9} & \textbf{59.5} & \textbf{42.7} & \textbf{39.7} & \textbf{24.9} \\
        \bottomrule
    \end{tabular}
    \label{tab:ablation_clustering}
\end{table*}


\noindent \textbf{Effects of Number of Key Steps.}
To understand the effect of the number of key steps, we evaluate RGWOT with varying $K$ values, as shown in Table~\ref{tab:k_ablation}. The model achieves its highest performance at $K=7$. Increasing $K$ beyond this leads to a consistent decline, with performance falling off significantly between 7 and 10. This trend holds across both datasets, indicating that $K=7$ is optimal for our model. The average number of distinct key steps across all datasets is 7, hence it is the optimal number of clusters.
\begin{table}[t]
    \centering
    \small
    \caption{Effects of number of key steps $K$ on performance across PC Assembly and PC Disassembly. Best results are in \textbf{bold}, while second best are \underline{underlined}.}
    \setlength{\tabcolsep}{8.7pt}
    \begin{tabular}{c ccc ccc}
        \toprule
        \multirow{2.5}{*}{$K$} & \multicolumn{3}{c}{PC Assembly} & \multicolumn{3}{c}{PC Disassembly} \\
        \cmidrule(lr){2-4} \cmidrule(lr){5-7}
        & R & F1 & IoU & R & F1 & IoU \\
        \midrule
        7  & \textbf{48.0} & \textbf{43.6} & \textbf{28.0} & \textbf{48.7} & \textbf{45.9} & \textbf{30.1} \\
        10 & \underline{37.8} & \underline{37.8} & \underline{23.5} & \underline{36.7} & \underline{36.6} & \underline{23.7} \\
        12 & 36.2 & 36.5 & 22.4 & 35.8 & 35.5 & 22.9 \\
        15 & 35.1 & 35.9 & 22.0 & 34.9 & 34.7 & 21.4 \\
        \bottomrule
    \end{tabular}
    \label{tab:k_ablation}
\end{table}

\begin{figure}[t]
    \centering
    \includegraphics[width=0.475\textwidth]{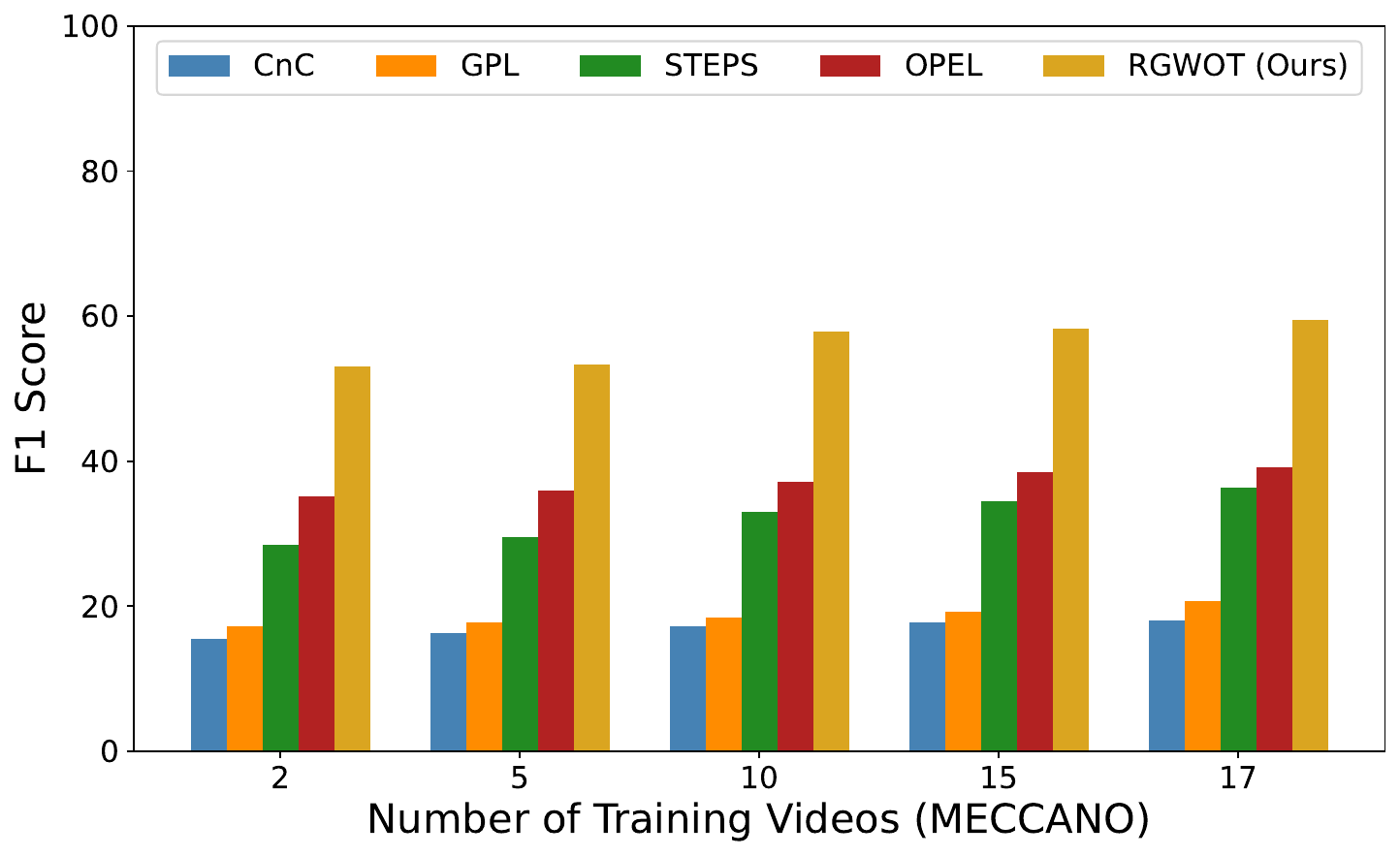} 
    \caption{Effects of training data quantity on F1 score across different methods on MECCANO~\cite{ragusa2021meccano}.}
    \label{fig:Training Data Quantity}
\end{figure}

\noindent \textbf{Effects of Training Data Quantity.}
Fig.~\ref{fig:Training Data Quantity} shows how model performance improves as more training videos are used on the MECCANO~\cite{ragusa2021meccano} dataset. Our method, RGWOT, consistently outperforms previous methods across all training set sizes. Even with just two training videos, RGWOT achieves a strong F1 score, surpassing the best scores of the other methods. As more data is added, its performance increases steadily, reaching 59.5\% with 17 videos. In contrast, previous methods exhibit slower gains and lower overall performance. These results highlight RGWOT’s superior data efficiency and scalability.

\noindent \textbf{Supplementary Material.} Due to space constraints, please refer to our supplementary material for our hyperparameter settings, sensitivity analyses, run time comparisons, degenerate solution justification, and quantitative results on all subtasks of EgoProceL~\cite{bansal2022my}, ProceL~\cite{elhamifar2019unsupervised}, and CrossTask~\cite{zhukov2019cross}.
\section{Conclusion}
\label{sec:conclusion}

We present a self-supervised procedure learning framework for recognizing key steps and their order from a collection of unlabeled procedural videos. In particular, our approach leverages a fused Gromov-Wasserstein optimal transport formulation with a structural prior to learn frame-to-frame correspondences between videos while tackling order variations, background/redundant frames, and repeated actions. Furthermore, we empirically examine possible trivial solutions when optimizing solely for the above temporal alignment term and integrate a contrastive regularization term which encourages various frames to be mapped to various points in the embedding space to prohibit the collapse to degenerate solutions. Lastly, extensive evaluations on large-scale egocentric (i.e., EgoProceL) and third-person (i.e., ProCeL and CrossTask) benchmarks are performed to show our superior performance over previous methods, including OPEL which is based on a classical Kantorovich optimal transport formulation with an optimality prior. Our future work will focus on deriving theoretical analyses of degenerate solutions,  utilizing alternative regularizations (e.g., TCLR~\cite{dave2022tclr}), and explore skeleton-based approaches~\cite{hyder2024action,tran2024learning}.

\appendix
\section{Supplementary Material}
In this supplementary material, we first provide hyperparameter settings of our RGWOT approach across all datasets in Sec.~\ref{sec:supp_hyperparameter_settings}. Next, we present sensitivity analyses of the margin $\lambda$ and weight $\beta$ of the C-IDM regularization, as well as the Gromov-Wasserstein weight $\alpha$, structural prior radius $r$, and temporal prior weight $\rho$ in Sec.~\ref{sec:supp_sensitivity_analysis}, followed by run time comparisons of our RGWOT approach and competing methods including VAOT~\cite{ali2025joint} and OPEL~\cite{chowdhuryopel} in Sec.~\ref{sec:supp_runtime_comparison}. We further provide an empirical justification of degenerate solutions in Sec.~\ref{sec:supp_degenerate_results}. Finally, we present quantitative results on all subtasks of egocentric and third-person datasets in Sec.~\ref{sec:supp_quantitative_results}.

\subsection{Hyperparameter Settings}
\label{sec:supp_hyperparameter_settings}

Implementation details have been provided in Sec.~4 of the main paper. Here, we additionally provide hyperparameter settings of our RGWOT approach across all datasets. Tab.~\ref{tab:supp_hyperparameter} lists  hyperparameter settings for RGWOT, including the learning rate, optimizer, window size, and other relevant hyperparameters. Note that we use the same hyperparameter settings across all datasets.

\begin{table}[t]
    \centering
    \caption{Hyperparameter settings for RGWOT.}
    \setlength{\tabcolsep}{5pt}
    \begin{tabular}{@{}lr@{}}
        \toprule
        Hyperparameter & Value \\
        \midrule
        No. of key-steps ($K$) & $7$\\
        No. of sampled frames ($X$,$Y$) & $32$\\
        No. of epochs & $10000$\\
        Batch Size & $2$\\
        Learning Rate ($\theta$) & $10^{-4}$\\
        Weight Decay & $10^{-5}$\\
        Window size ($\sigma$) & $300$\\
        Margin ($\lambda$) & $2.0$\\
        No. of context frames ($c$) & $2$\\
        Context stride & $15$\\
        Embedding Dimension ($D$) & $128$ \\
        Optimizer & Adam~\cite{kingma2014adam}\\
        Regularization parameter ($\xi$) & $1.0$\\
        Entropy regularization weight ($\epsilon$) & $0.07$\\
        Virtual frame threshold ($\zeta$) & $0.5$\\
        Gromov-Wasserstein weight ($\alpha$) & $0.3$\\
        Structural prior radius ($r$) & $0.02$\\
        Temporal prior weight ($\rho$) & $0.35$\\
        \bottomrule
    \end{tabular}
    \label{tab:supp_hyperparameter}
\end{table}

\subsection{Sensitivity Analyses}
\label{sec:supp_sensitivity_analysis}

\begin{figure*}[t]
    \centering
    \resizebox{0.98\textwidth}{!}{
    \begin{minipage}{\textwidth}
        \begin{subfigure}[b]{0.4\textwidth}
            \includegraphics[width=\textwidth]{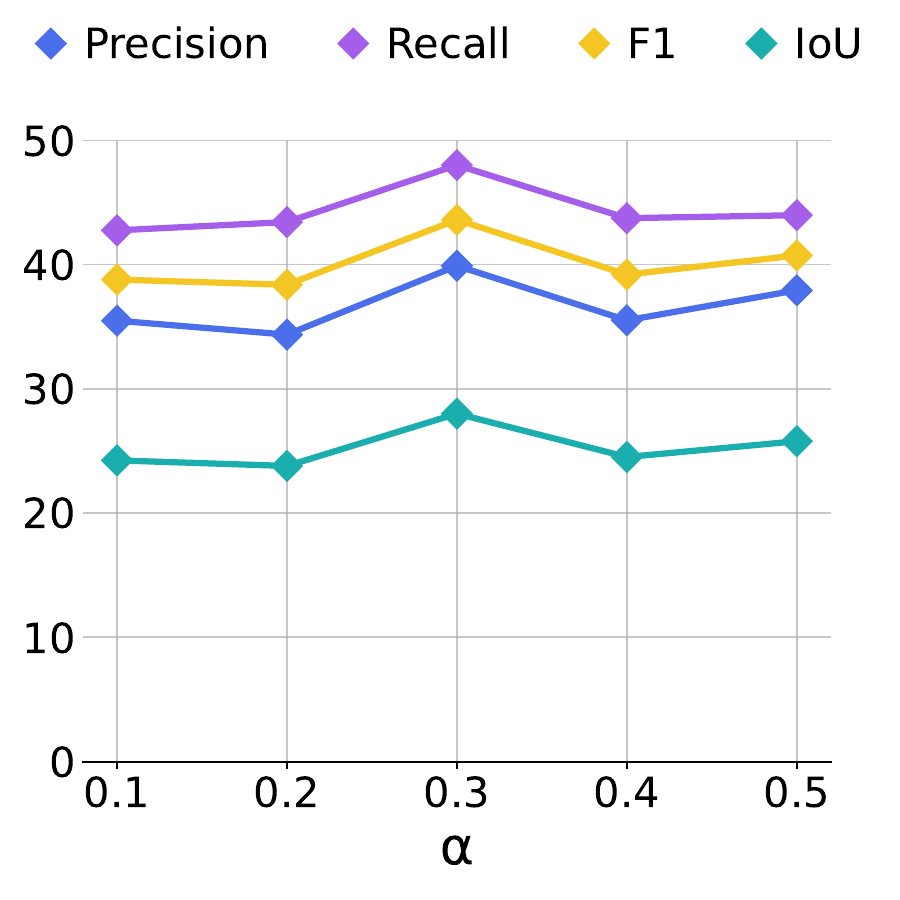}
            \caption{Gromov-Wasserstein weight $\alpha$.}
        \end{subfigure}\hfill
        \begin{subfigure}[b]{0.4\textwidth}
            \includegraphics[width=\textwidth]{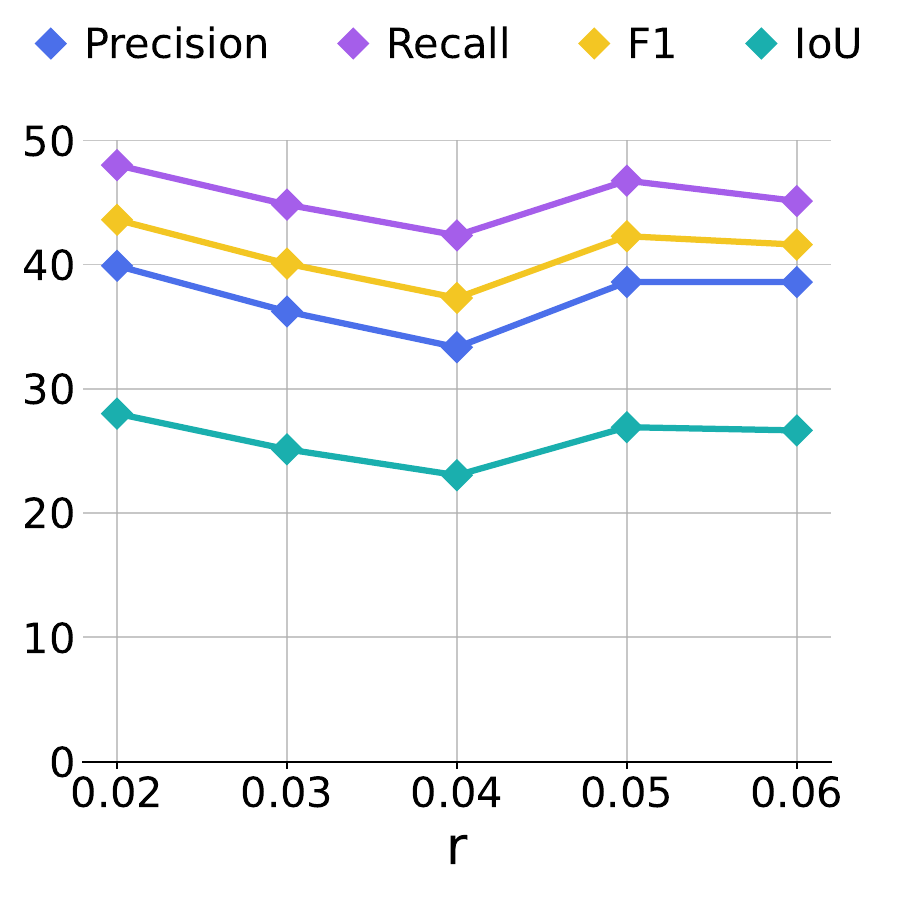}
            \caption{Structural prior radius $r$.}
        \end{subfigure}

        \begin{subfigure}[b]{0.4\textwidth}
            \includegraphics[width=\textwidth]{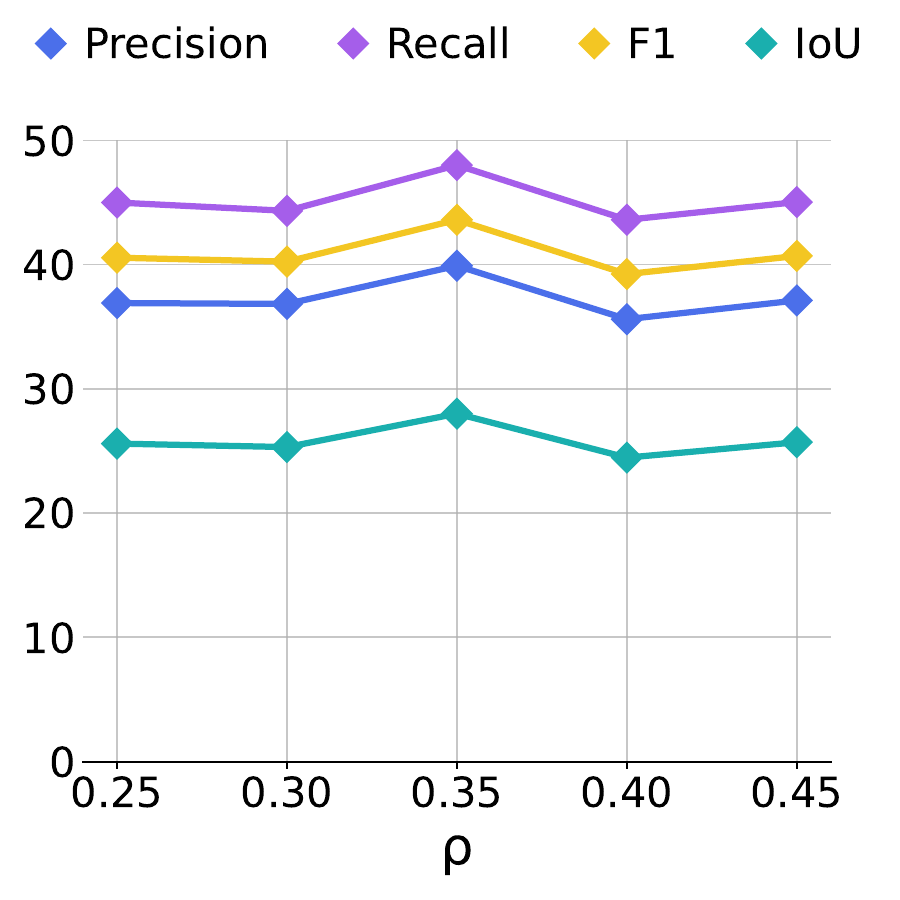}
            \caption{Temporal prior weight $\rho$.}
        \end{subfigure}\hfill
        \begin{subfigure}[b]{0.4\textwidth}
            \includegraphics[width=\textwidth]{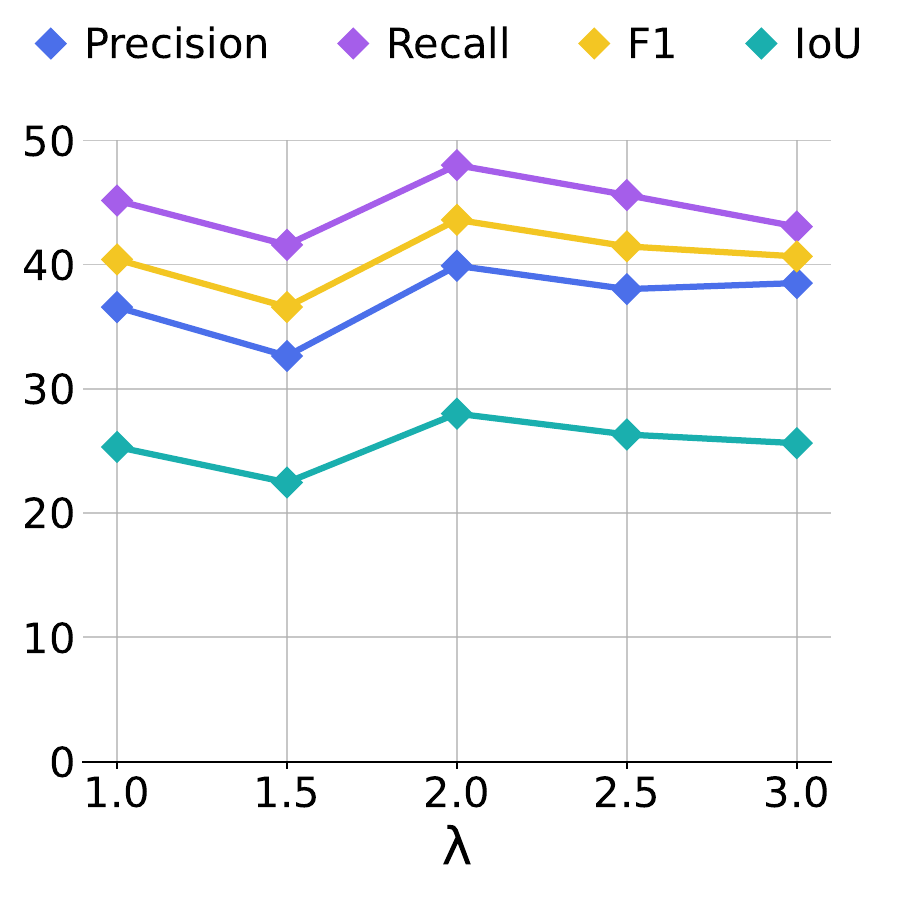}
            \caption{Contrastive regularization margin $\lambda$.}
        \end{subfigure}
    \end{minipage}
    }
    \caption{Sensitivity analyses of Gromov-Wasserstein weight $\alpha$, structural prior radius $r$, temporal prior weight $\rho$, and contrastive regularization margin $\lambda$.}
    \label{fig:sens}
\end{figure*}

In addition to the sensitivity analysis of $K$ in Sec.~4.2 of the main paper, we further conduct sensitivity analyses on some key hyperparameters of our RGWOT approach, including the Gromov–Wasserstein weight $\alpha$, structural prior radius $r$, temporal prior weight $\rho$, as well as the margin $\lambda$ and weight $\beta$ of the C-IDM regularization. First of all, Fig.~\ref{fig:sens} presents sensitivity analyses of the Gromov–Wasserstein weight $\alpha$, structural prior radius $r$, temporal prior weight $\rho$, and margin $\lambda$ on the EgoProceL~\cite{bansal2022my} dataset (i.e., PC Assembly). Each subfigure illustrates how the Precision, Recall, F1, and IoU metrics change as the corresponding hyperparameter is adjusted. Overall, RGWOT demonstrates mostly stable performance across the studied ranges, with the performance metrics reaching their peaks at $\alpha=0.3$, $r=0.02$, $\rho=0.35$, and $\lambda=2$.

Next, we study the effects of the C-IDM regularization by varying its weight $\beta$ in Eq.~7 of the main paper and include the results on the ProceL~\cite{elhamifar2020self} dataset (i.e., make\_smoke\_salmon\_sandwich) in Tab.~\ref{tab:supp_degenerate_results}. It is evident from the results that when $\beta$ is small (e.g., $\beta=0.01$), the learning collapses and RGWOT yields degenerate solutions, where all frames are mapped to a small cluster in the embedding space, and hence an entire video is assigned to a single key steps (see Fig.~3 for examples). Furthermore, when $\beta$ increases (e.g., $\beta=0.05$), the above issue is prevented. Finally, the best performance is achieved when $\beta=1$, indicating balancing weights for the video alignment loss and the contrastive regularization.

\begin{table}[t]
\centering
\small
\begin{tabular}{ccccc}
\hline
\textbf{$\beta$} & \textbf{Precision} & \textbf{Recall} & \textbf{F1} & \textbf{IoU} \\
\hline
0.01  & X & X & X & X \\
0.05  & 38.51 & 40.34 & 39.40 & 24.02 \\
0.1  & \underline{40.49} & 42.68 & \underline{41.56} & \underline{26.73} \\
1    & \textbf{42.61} & \underline{44.73} & \textbf{43.64} & \textbf{28.22} \\
10   & 36.41 & \textbf{44.87} & 40.20 & 25.33 \\
50   & 35.41 & 40.97 & 37.99 & 23.54 \\
\hline
\end{tabular}
\caption{Sensitivity analysis of the contrastive regularization weight $\beta$. Best results are in \textbf{bold}, while second best are \underline{underlined}. `X' denotes degenerate results.}
\label{tab:supp_degenerate_results}
\end{table}

\subsection{Run Time Comparisons}
\label{sec:supp_runtime_comparison}

In this section, we evaluate the efficiency (in terms of training times) of our RGWOT approach and competing methods including OPEL~\cite{chowdhuryopel} and VAOT~\cite{ali2025joint}. In particular, we train all methods on the EgoProceL~\cite{bansal2022my} dataset (i.e., PC Disassembly) under identical experimental conditions (e.g., with 10,000 training epochs using two NVIDIA RTX 5090 GPUs). Tab.~\ref{tab:training_times} presents the results. Thanks to having few losses and regularizers, our RGWOT approach is (notably) more efficient than OPEL~\cite{chowdhuryopel}, i.e., for training, OPEL~\cite{chowdhuryopel} needs $206$ minutes vs. $161$ minutes for RGWOT. Thus, our approach is not only more accurate but also more efficient than OPEL~\cite{chowdhuryopel}. In addition, RGWOT has similar training time as VAOT~\cite{ali2025joint}, i.e., for training, VAOT~\cite{ali2025joint} requires $164$ minutes vs. $161$ minutes for RGWOT, suggesting that adding contrastive regularization yields little extra computation.

\begin{table}[t]
    \centering
    \caption{Run time comparisons.}
    \setlength{\tabcolsep}{12pt}
    \begin{tabular}{@{}cc@{}}
        \toprule
        Method & Training Time \\
        \midrule
        OPEL~\cite{chowdhuryopel} & 206 (mins) \\
        VAOT~\cite{ali2025joint} & 161 (mins) \\
        RGWOT (Ours) & 164 (mins) \\
        \bottomrule
    \end{tabular}
    \label{tab:training_times}
\end{table}

\subsection{Degenerate Solution Justification}
\label{sec:supp_degenerate_results}

\begin{figure*}[!htbp]
    \centering

    \begin{subfigure}[b]{0.27\textwidth}
        \centering
        \includegraphics[width=\textwidth]{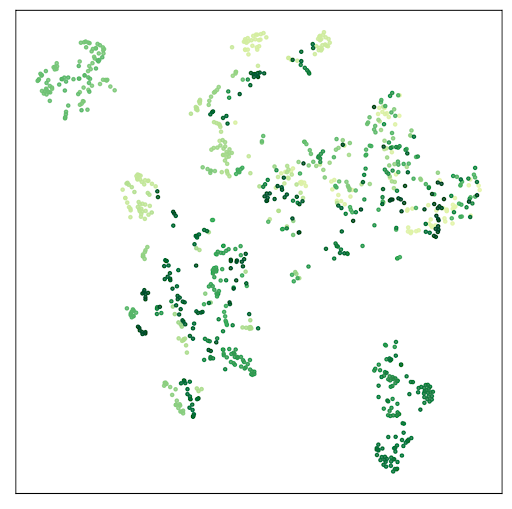}
        \caption{}
        \label{fig:tsne-egoprocel1}
    \end{subfigure}
    \begin{subfigure}[b]{0.27\textwidth}
        \centering
        \includegraphics[width=\textwidth]{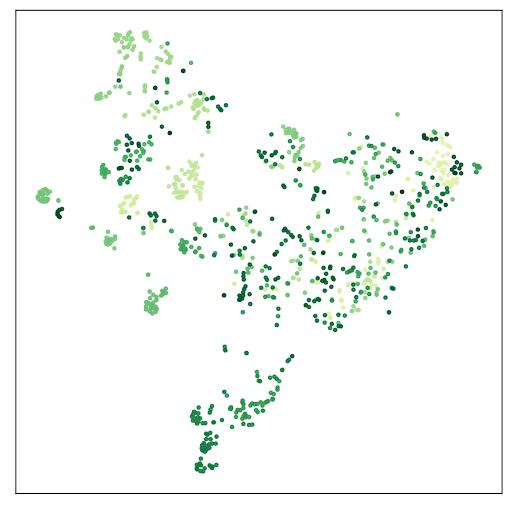}
        \caption{}
        \label{fig:tsne-egoprocel2}
    \end{subfigure}

    \begin{subfigure}[b]{0.27\textwidth}
        \centering
        \includegraphics[width=\textwidth]{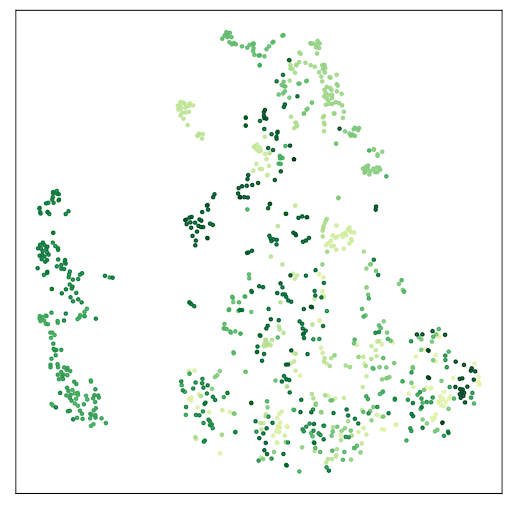}
        \caption{}
        \label{fig:tsne-egoprocel3}
    \end{subfigure}
    \begin{subfigure}[b]{0.27\textwidth}
        \centering
        \includegraphics[width=\textwidth]{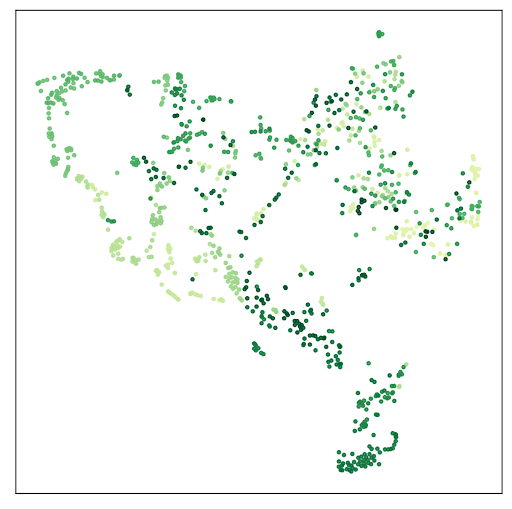}
        \caption{}
        \label{fig:tsne-egoprocel4}
    \end{subfigure}
    
    \caption{t-SNE~\cite{maaten2008visualizing} plots of the EgoProceL~\cite{bansal2022my} dataset. Frame order from first to last is indicated by a gradient from light to dark green.}
    \label{fig:tsne-egoprocel}
\end{figure*}

\begin{figure*}[!htbp]
    \centering
    
    \begin{subfigure}[b]{0.27\textwidth}
        \centering
        \includegraphics[width=\textwidth]{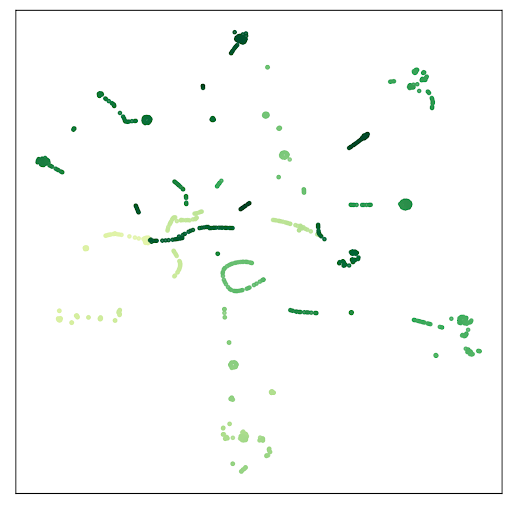}
        \caption{}
        \label{fig:tsne-procel1}
    \end{subfigure}
    \begin{subfigure}[b]{0.27\textwidth}
        \centering
        \includegraphics[width=\textwidth]{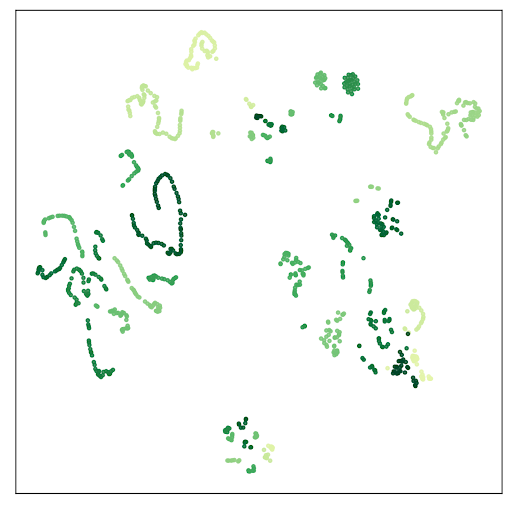}
        \caption{}
        \label{fig:tsne-procel2}
    \end{subfigure}
    
    \begin{subfigure}[b]{0.27\textwidth}
        \centering
        \includegraphics[width=\textwidth]{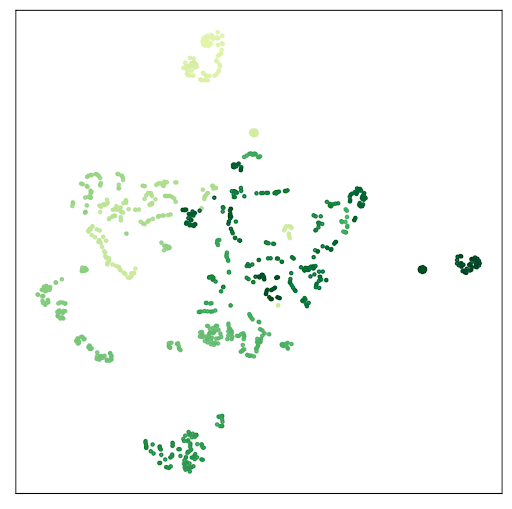}
        \caption{}
        \label{fig:tsne-procel3}
    \end{subfigure}
    \begin{subfigure}[b]{0.27\textwidth}
        \centering
        \includegraphics[width=\textwidth]{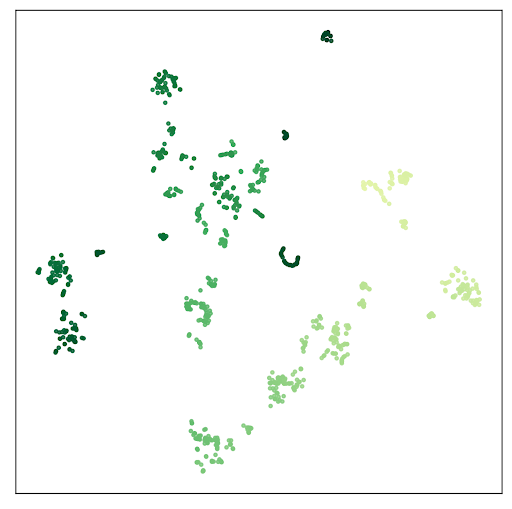}
        \caption{}
        \label{fig:tsne-procel4}
    \end{subfigure}
    
    \caption{t-SNE~\cite{maaten2008visualizing} plots of the ProceL~\cite{elhamifar2020self} dataset. Frame order from first to last is indicated by a gradient from light to dark green.}
    \label{fig:tsne-procel}
\end{figure*}

As discussed in Sec.~3.1.2 of the main paper, the objective in Eq.~5 minimizes visual and temporal differences between corresponding frames in $X$ and $Y$ only, and thus there is no mechanism to prevent the optimization from collapsing. Degenerate solutions appear consistently on the ProceL~\cite{elhamifar2020self} dataset but not on the EgoProceL~\cite{bansal2022my} and CrossTask~\cite{zhukov2019cross} datasets. It is likely because 1) ADAM is a local optimizer, and 2) datasets exhibit different characteristics. To verify the latter, we extract ResNet~\cite{he2016deep} features (which are input to the model) from four EgoProceL videos and four ProceL videos and visualize them using t-SNE~\cite{maaten2008visualizing} in Figs.~\ref{fig:tsne-egoprocel} and \ref{fig:tsne-procel} respectively. We observe that EgoProceL features are more spread out, whereas ProceL features are more concentrated and hence easier to collapse. We believe this empirical evidence provides a convincing explanation and consider a theoretical analysis an interesting avenue for our future work.

\subsection{Quantitative Results on All Subtasks of Egocentric and Third-Person Datasets}
\label{sec:supp_quantitative_results}

Tabs.~\ref{tab:egtea_subtask} and~\ref{tab:cmu_subtask} present quantitative results on all subtasks of egocentric datasets, namely EGTEA-GAZE+~\cite{li2018eye} and CMU-MMAC~\cite{de2009guide} respectively. Corresponding results for third-person datasets, including ProceL~\cite{elhamifar2020self} and CrossTask~\cite{zhukov2019cross}, are provided in Tabs.~\ref{tab:supp_procel_subtasks} and~\ref{tab:crosstask_subtask} respectively. Our detailed evaluations span a wide range of scenarios, offering detailed assessments of our model performance from different viewpoints. The results highlight the robustness and versatility of our approach in addressing diverse videos and tasks, contributing to progress in procedure learning and related fields.

\begin{table*}[t]
    \centering
    \caption{Results on individual subtasks of the EGTEA-GAZE+~\cite{li2018eye} dataset.}
    \setlength{\tabcolsep}{12pt}
    \begin{tabular}{cccccccccc}
        \toprule
        & Bacon Eggs & Cheeseburger & Breakfast & Greek Salad & Pasta Salad & Pizza & Turkey \\
        \midrule
        F1  & 37.8 & 41.2 & 34.5 & 37.3 & 35.0 & 37.1 & 38.7\\
        IoU & 23.4 & 26.2 & 21.0 & 23.0 & 21.4 & 22.9 & 22.8\\
        \bottomrule
    \end{tabular}
    \label{tab:egtea_subtask}
\end{table*}

\begin{table*}[t]
    \centering
    \caption{Results on individual subtasks of the CMU-MMAC~\cite{de2009guide} dataset.}
    \setlength{\tabcolsep}{28pt}
    \begin{tabular}{cccccccccc}
        \toprule
        & Brownie & Eggs & Sandwich & Salad & Pizza \\
        \midrule
        F1  & 59.5 & 44.1 & 56.1 & 68.6 & 43.7\\
        IoU & 42.8 & 28.5 & 39.9 & 53.3 & 28.2\\
        \bottomrule
    \end{tabular}
    \label{tab:cmu_subtask}
\end{table*}

\begin{table*}[t]
    \centering
    \caption{Results on individual subtasks of the ProceL~\cite{elhamifar2020self} dataset.}
    \setlength{\tabcolsep}{4.9pt}
    \begin{tabular}{ccccccccc}
        \toprule
        & Assemble Clarinet & Change iPhone Battery & Change Tire & Change Toilet Seat & Jump Car & Coffee \\
        \midrule
        F1  & 56.4 & 48.8 & 42.7 & 48.2 & 35.2 & 48.8 \\
        IoU & 41.6 & 33.5 & 27.6 & 32.5 & 21.7 & 33.3 \\
        \midrule
        & Make PBJ Sandwich & Make Salmon Sandwich & Perform CPR & Repot Plant & Setup Chromecast & Tie Tie \\
        \midrule
        F1  & 38.8 & 43.6 & 42.1 & 47.4 & 36.6 & 43.7 \\
        IoU & 24.3 & 28.2 & 27.1 & 32.2 & 22.7 & 28.1 \\
        \bottomrule
    \end{tabular}
    \label{tab:supp_procel_subtasks}
\end{table*}

\begin{table*}[t]
    \centering
    \caption{Results on individual subtasks of the CrossTask~\cite{zhukov2019cross} dataset.}
    \setlength{\tabcolsep}{11.8pt}
    \begin{tabular}{cccccccccc}
        \toprule
        & 40567 & 16815 & 23521 & 44047 & 44789 & 77721 & 87706 & 71781 & 94276 \\
        \midrule
        F1  & 42.8 & 53.8 & 33.9 & 34.7 & 31.7 & 38.0 & 33.1 & 32.8 & 39.5 \\
        IoU & 27.2 & 37.4  & 20.6 & 21.2  & 18.9  & 23.7  & 20.3 & 19.8  & 25.1 \\
        \midrule
        & 53193 & 76400 & 91515 & 59684 & 95603 & 105253 & 105222 & 109972 & 113766 \\
        \midrule
        F1  & 38.5 & 34.4 & 52.3 & 36.1 & 53.8 & 52.0 & 40.4 & 42.8 & 37.4 \\
        IoU & 24.1 & 20.9 & 37.1 & 23.6 & 38.4 & 37.6 & 26.3 & 28.0 & 23.4 \\
        \bottomrule
    \end{tabular}
    \label{tab:crosstask_subtask}
\end{table*}

{
\small
\bibliographystyle{ieeenat_fullname}
\bibliography{references}
}

\end{document}